\documentclass{vgtc}                          

\ifpdf
  \pdfoutput=1\relax                   
  \usepackage{graphicx}                
  \DeclareGraphicsExtensions{.pdf,.png,.jpg,.jpeg} 
\else
  \usepackage{graphicx}                
  \DeclareGraphicsExtensions{.eps}     
\fi%

\graphicspath{{figures/}{pictures/}{images/}{./}} 

\usepackage{microtype}                 
\PassOptionsToPackage{warn}{textcomp}  
\usepackage{textcomp}                  
\usepackage{mathptmx}                  
\usepackage{times}                     
\usepackage{cite}                      
\usepackage{tabu}                      
\usepackage{booktabs}                  
\usepackage{amsmath}
\usepackage[nolist]{acronym}
\usepackage{amsmath}
\usepackage{listings}
\usepackage[table,xcdraw]{xcolor}
\usepackage{xcolor}
\definecolor{maroon}{cmyk}{0, 0.87, 0.68, 0.32}
\definecolor{halfgray}{gray}{0.55}
\definecolor{ipython_frame}{RGB}{207, 207, 207}
\definecolor{ipython_bg}{RGB}{247, 247, 247}
\definecolor{ipython_red}{RGB}{186, 33, 33}
\definecolor{ipython_green}{RGB}{0, 128, 0}
\definecolor{ipython_cyan}{RGB}{64, 128, 128}
\definecolor{ipython_purple}{RGB}{170, 34, 255}

\usepackage{listings}
\lstdefinelanguage{iPython}{
    morekeywords={access,and,break,class,continue,def,del,elif,else,except,exec,finally,for,from,global,if,import,in,is,lambda,not,or,pass,print,raise,return,try,while},%
    morekeywords=[2]{abs,all,any,basestring,bin,bool,bytearray,callable,chr,classmethod,cmp,compile,complex,delattr,dict,dir,divmod,enumerate,eval,execfile,file,filter,float,format,frozenset,getattr,globals,hasattr,hash,help,hex,id,input,int,isinstance,issubclass,iter,len,list,locals,long,map,max,memoryview,min,next,object,oct,open,ord,pow,property,range,raw_input,reduce,reload,repr,reversed,round,set,setattr,slice,sorted,staticmethod,str,sum,super,tuple,type,unichr,unicode,vars,xrange,zip,apply,buffer,coerce,intern},%
    sensitive=true,%
    morecomment=[l]\#,%
    morestring=[b]',%
    morestring=[b]",%
    morestring=[s]{'''}{'''},
    morestring=[s]{"""}{"""},
    morestring=[s]{r'}{'},
    morestring=[s]{r"}{"},%
    morestring=[s]{r'''}{'''},%
    morestring=[s]{r"""}{"""},%
    morestring=[s]{u'}{'},
    morestring=[s]{u"}{"},%
    morestring=[s]{u'''}{'''},%
    morestring=[s]{u"""}{"""},%
    literate=
    {á}{{\'a}}1 {é}{{\'e}}1 {í}{{\'i}}1 {ó}{{\'o}}1 {ú}{{\'u}}1
    {Á}{{\'A}}1 {É}{{\'E}}1 {Í}{{\'I}}1 {Ó}{{\'O}}1 {Ú}{{\'U}}1
    {à}{{\`a}}1 {è}{{\`e}}1 {ì}{{\`i}}1 {ò}{{\`o}}1 {ù}{{\`u}}1
    {À}{{\`A}}1 {È}{{\'E}}1 {Ì}{{\`I}}1 {Ò}{{\`O}}1 {Ù}{{\`U}}1
    {ä}{{\"a}}1 {ë}{{\"e}}1 {ï}{{\"i}}1 {ö}{{\"o}}1 {ü}{{\"u}}1
    {Ä}{{\"A}}1 {Ë}{{\"E}}1 {Ï}{{\"I}}1 {Ö}{{\"O}}1 {Ü}{{\"U}}1
    {â}{{\^a}}1 {ê}{{\^e}}1 {î}{{\^i}}1 {ô}{{\^o}}1 {û}{{\^u}}1
    {Â}{{\^A}}1 {Ê}{{\^E}}1 {Î}{{\^I}}1 {Ô}{{\^O}}1 {Û}{{\^U}}1
    {œ}{{\oe}}1 {Œ}{{\OE}}1 {æ}{{\ae}}1 {Æ}{{\AE}}1 {ß}{{\ss}}1
    {ç}{{\c c}}1 {Ç}{{\c C}}1 {ø}{{\o}}1 {å}{{\r a}}1 {Å}{{\r A}}1
    {€}{{\EUR}}1 {£}{{\pounds}}1,
    literate=
    *{+}{{{\color{ipython_purple}+}}}1
    {-}{{{\color{ipython_purple}-}}}1
    {*}{{{\color{ipython_purple}$^\ast$}}}1
    {/}{{{\color{ipython_purple}/}}}1
    {^}{{{\color{ipython_purple}\^{}}}}1
    {?}{{{\color{ipython_purple}?}}}1
    {!}{{{\color{ipython_purple}!}}}1
    {\%}{{{\color{ipython_purple}\%}}}1
    {<}{{{\color{ipython_purple}<}}}1
    {>}{{{\color{ipython_purple}>}}}1
    {|}{{{\color{ipython_purple}|}}}1
    {\&}{{{\color{ipython_purple}\&}}}1
    {~}{{{\color{ipython_purple}~}}}1
    {==}{{{\color{ipython_purple}==}}}2
    {<=}{{{\color{ipython_purple}<=}}}2
    {>=}{{{\color{ipython_purple}>=}}}2
    {+=}{{{+=}}}2
    {-=}{{{-=}}}2
    {*=}{{{$^\ast$=}}}2
    {/=}{{{/=}}}2,
    commentstyle=\color{ipython_cyan}\ttfamily,
    stringstyle=\color{ipython_red}\ttfamily,
    keepspaces=true,
    showspaces=false,
    showstringspaces=false,
    rulecolor=\color{ipython_frame},
    frame=single,
    frameround={t}{t}{t}{t},
    framexleftmargin=0mm,
    numbers=left,
    numberstyle=\tiny\color{halfgray},
    backgroundcolor=\color{ipython_bg},
    basicstyle=\scriptsize\ttfamily,
    keywordstyle=\color{ipython_green}\ttfamily,
    escapechar=\¢,escapebegin=\color{ipython_green},
}
\usepackage{booktabs}
\usepackage{listings}
\usepackage{array}
\usepackage{multirow}
\usepackage{steinmetz}
\usepackage{amsmath}
\usepackage[final]{pdfpages}

\usepackage{booktabs} 
\usepackage{xspace}
\newcommand{\etal}{~et al.\@\xspace} 
\newcommand{\Fig}{Fig.\@\xspace} 
\usepackage{color}
\usepackage{xcolor}
\usepackage{savesym} \savesymbol{spacing} 
\usepackage{setspace}
\usepackage{tikz}
\usepackage{colortbl}
\definecolor{Orange}{rgb}{1,0.5,0}
\definecolor{DarkGreen}{rgb}{0,0.5,0}
\definecolor{Purple}{rgb}{0.7,0,0.7}
\definecolor{Blue}{rgb}{0.2,0.2,0.8}
\definecolor{Red}{rgb}{1.0,0.0,0.0}
\definecolor{Brown}{rgb}{0.7,0.4,0.1}
\definecolor{Blue}{rgb}{0, 0, 1.}
\definecolor{Green}{rgb}{0., .6, 0.}
\definecolor{Custom}{rgb}{0.3, .1, 0.2}
\definecolor{Yellow}{rgb}{.9, .7, 0.}
\definecolor{Purple}{rgb}{.9, .1, 0.8}

\newcommand{\NewEmptyPage}{\newpage\null\thispagestyle{empty}\newpage}

\onlineid{1042}

\vgtccategory{Research}

\vgtcinsertpkg

\title{Realistic Defocus Blur for Multiplane Computer-Generated Holography}

\author{Koray Kavakl{\i}\thanks{e-mail: kkavakli@ku.edu.tr}\\ %
     \parbox{1.4in}{\scriptsize \centering Ko\c{c} University \\ University College London} %
\and Yuta Itoh\thanks{e-mail: yuta.itoh@iii.u-tokyo.ac.jp}\\ %
     \scriptsize The  University of Tokyo %
\and Hakan Urey\thanks{e-mail: hurey@ku.edu.tr}\\ %
     \scriptsize Ko\c{c} University %
\and Kaan Ak\c{s}it\thanks{e-mail: k.aksit@ucl.ac.uk}\\ %
     \scriptsize University College London}

\teaser{
  \centering
  \includegraphics[width=\linewidth]{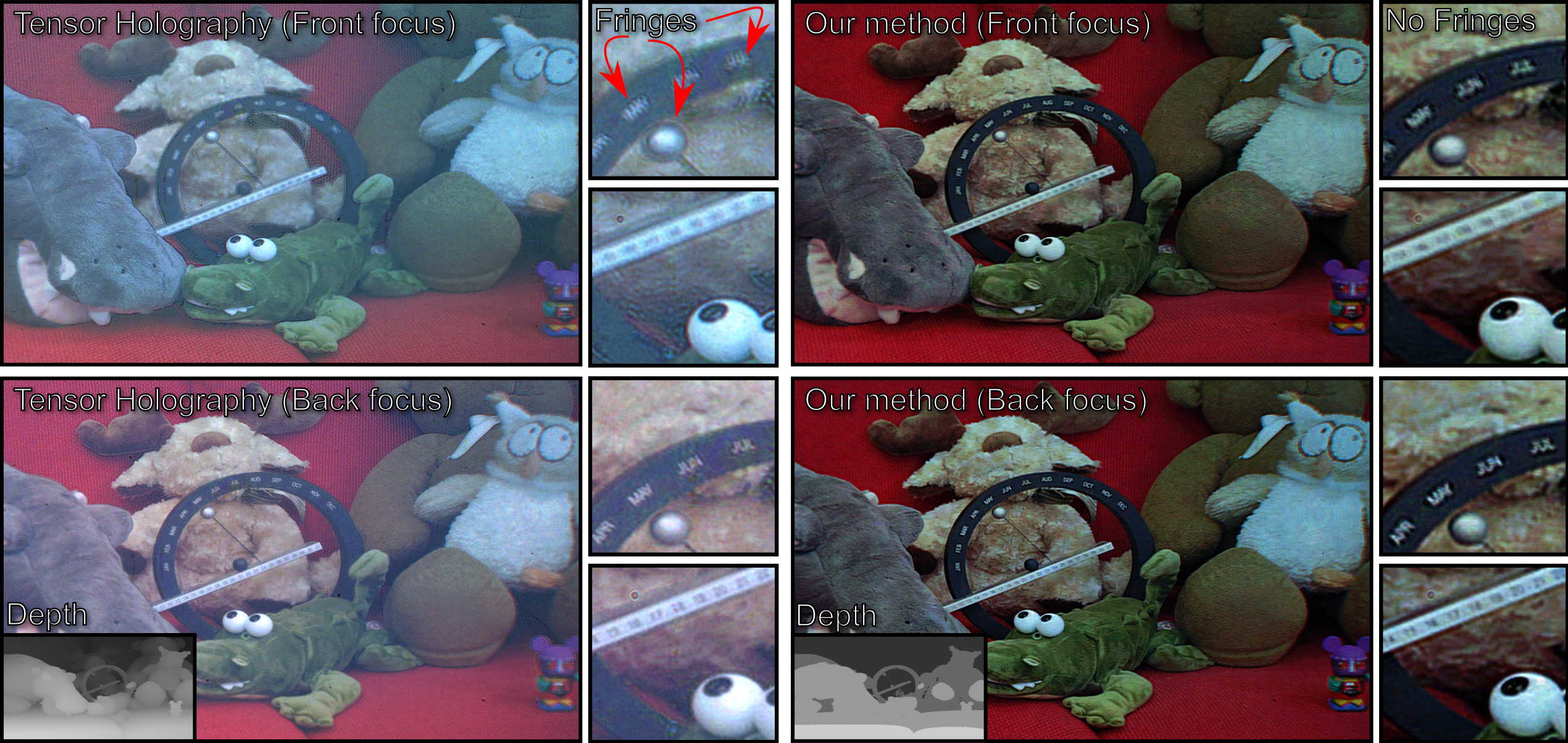}
  \caption{Realistic defocus blur in multiplane Computer-Generated Holography (CGH).
We introduce a novel hologram generation pipeline for holographic displays that offer near-accurate defocus blur and mitigation of visible fringes in reconstructed images.
Here, we provide actual photographs captured from our proof-of-concept holographic display.
We capture three components of a color image using a three-color laser light source (473-515-639~nm) using the same capture exposure (20~ms) and laser powers, providing the best visual quality for each case.
We later combine three monochrome images into a full-color image in the post-process.
The state of the art method, tensor holography~\cite{shi2021towards}, offers sharp, high-quality images but with contrast loss.
Our proposed method mitigates these issues while offering 3D images with improved contrast, more realistic defocus blur and fringe-free images.
In this example, our approach here uses six target planes in-depth.
We produced the results for tensor holography~\cite{shi2021towards} with our hardware to the best of our knowledge.}
  \label{fig:teaser}
}

\abstract{This paper introduces a new multiplane CGH computation method to reconstruct artifact-free high-quality holograms with natural-looking defocus blur. 
Our method introduces a new targeting scheme and a new loss function.
While the targeting scheme accounts for defocused parts of the scene at each depth plane, the new loss function analyzes focused and defocused parts separately in reconstructed images.
Our method support phase-only CGH calculations using various iterative (e.g., Gerchberg-Saxton, Gradient Descent) and non-iterative (e.g., Double Phase) CGH techniques. 
We achieve our best image quality using a modified gradient descent-based optimization recipe where we introduce a constraint inspired by the double phase method.
We validate our method experimentally using our proof-of-concept holographic display, comparing various algorithms, including multi-depth scenes with sparse and dense contents.%
} 

\CCScatlist{
  \CCScatTwelve{Hardware}{Emerging Technologies}{Emerging optical and photonic technology}{};
  \CCScatTwelve{Hardware}{Communication hardware, interfaces and storage}{Display and imagers}{}
}

\begin{document}


\firstsection{Introduction}
\label{sec:introduction}

\maketitle
Real-time computer-generated visuals are confined chiefly to flat images.
Rendering realistic-looking perspectives of a scene on a flat two-dimensional screen is possible today. 
Nevertheless, the actual illusion of perceiving 3D scenes that match our experiences in real life is still beyond today's computer graphics and displays.
There is a growing consensus in the industry and academia that Computer-Generated Holography (CGH)~\cite{slinger2005computer} methods can help address these issues, and holography can stand out as the next-generation display technology~\cite{lucente1997interactive}.
\textit{The goal of CGH research is to provide computer-generated 3D visuals indistinguishable from real life by the human visual system.}

\begin{figure}[t]
\centering
\includegraphics[width=1.0\linewidth]{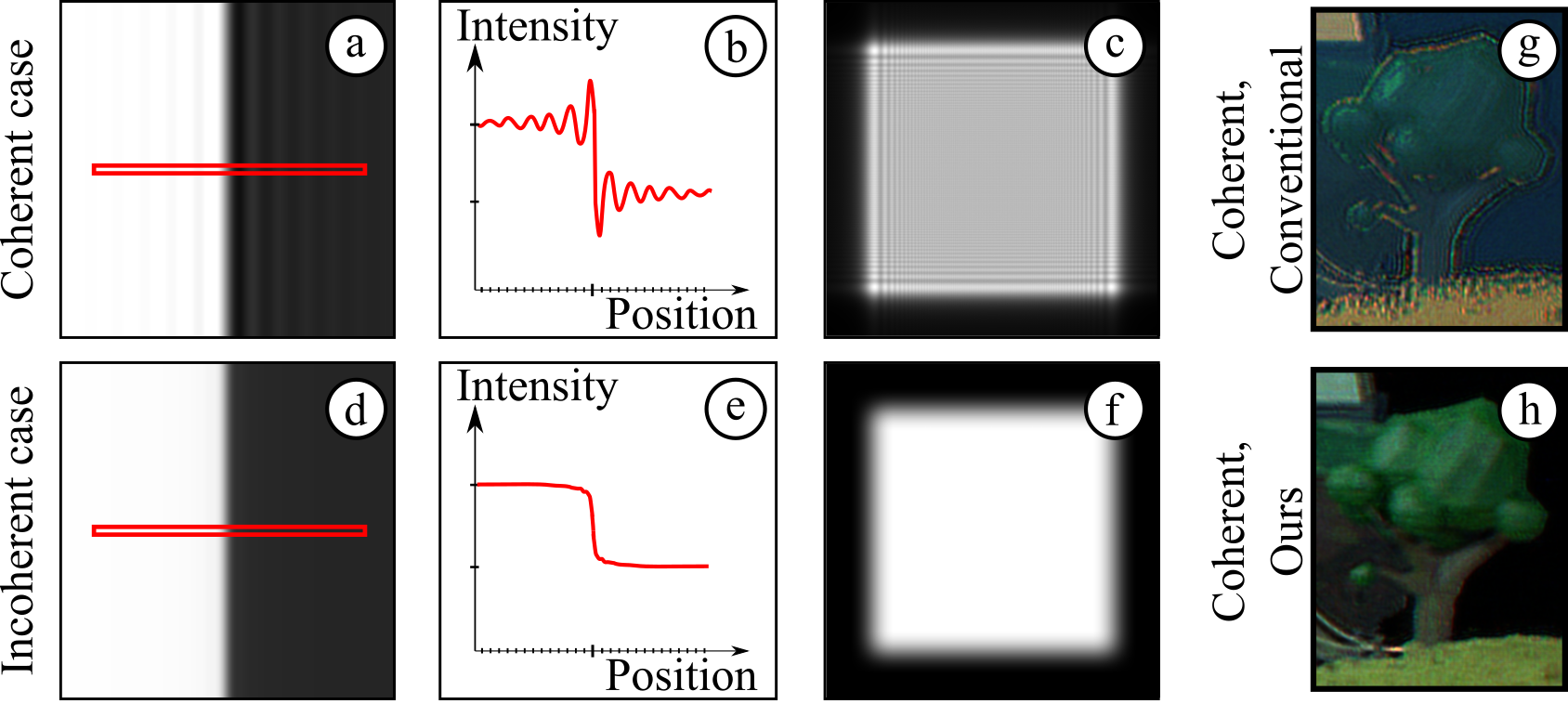}
\caption{
Differences in defocus blur when using coherent and incoherent illumination. 
According to Born and Wolf~\cite{born2013principles}, fringes are typical in 3D holographic displays at the defocused parts when using coherent illumination (a, b), whereas incoherent illumination (d, e) case does not suffer from such issues. 
Thus, coherent (c) and incoherent (f) illumination often differ in defocus blur visually. (g) Real captures of a conventional coherent hologram with fringes, and (h) that of our coherent hologram where fringes are suppressed.
}
\label{fig:edge_fringe_coherent_blur}
 \end{figure}

With the advent of machine learning techniques and accelerated computing capabilities, the image quality of visuals generated by CGH has improved tremendously, specifically for 3D case~\cite{shi2021towards}.
However, there are still barriers to achieving realistic visuals with CGH, specifically in the defocused parts of a scene.
As CGH typically relies on coherent light sources, it makes defocused parts of a scene look unfamiliar to a human observer.
Born and Wolf~\cite{born2013principles} describe this apparent \textit{difference between coherent and incoherent defocus blur} as the \textit{edge fringe} issue in 3D holography (see Figure~\ref{fig:edge_fringe_coherent_blur}).

In this work, we address the difference between coherent and incoherent blur in the context of multiplane CGH. 
We argue that an improved targeting scheme and a new loss function that accounts for these differences can help reproduce incoherent defocus blur in CGH when reconstructing multiplanar images using coherent light sources.
We show that such a loss function and a targeting scheme work with the most common CGH calculation methods, including Gerchberg-Saxton (GS)~\cite{zhou201530}, Stochastic Gradient Descent (SGD)~\cite{zhang20173d} and Double Phase (DP)~\cite{hsueh1978computer} based approaches.
We build a proof-of-concept holographic display to validate our method experimentally.
Our primary technical contributions include the following:

\begin{itemize}

\item \textbf{Loss function and targeting scheme.}
We introduce a new loss function that evaluates focused and defocused parts of target images in multiplane CGH.
We also present a new targeting scheme to set proper target images, including defocused parts for multiplane image reconstructions.
Our loss function and targeting scheme can generate near-accurate defocus blur approximating incoherent cases using coherent light.

\item \textbf{Multiplane hologram generation pipeline.}
Our loss function and targeting scheme are compatible with various optimization and learning methods.
We describe our hologram generation pipeline for reconstructing multiplane images on an SLM plane (0~cm away) or a near field plane (15~cm away).
Our hologram generation pipeline uses a custom optimization recipe that combines double phase constraints with SGD optimizations.

\item \textbf{Bench-top holographic display prototype.}
We implement a proof-of-concept holographic display prototype using three lasers, and a phase-only Spatial Light Modulator (SLM) augmented with various optical and optomechanical parts.
This prototype serves as a base for validating our algorithms for our method.

\end{itemize}

\section{Related Work}
\label{sec:related}
We introduce a new targeting scheme and a loss function to improve the visual quality of image reconstructions simultaneously at multiple planes -- not varifocal CGH~\cite{9756777}. 
Notably, our work deals with the edge fringe issues and differences in defocus blur between incoherent and coherent light in holographic displays. 
Here, we provide a brief survey of prior art in relevant CGH methods.

\subsection{Holographic Displays}
A holographic display~\cite{slinger2005computer} aims to produce genuine 3D light fields~\cite{lippmann1908epreuves} using the optical phenomena of diffraction and interference in coherent imaging~\cite{born2013principles}.
Typically, in these holographic displays, an SLM represents these diffraction and interference patterns in a programmable fashion.
While holographic displays can come in different types, such as near-eye displays~\cite{cem2020foveated}, desktop displays~\cite{Lee:16}, and even contact lens displays~\cite{Sano:21}, all these displays claim to offer near-correct optical focus cues~\cite{Zhang:15}.
However, the defocused parts represented in these displays do not look natural due to the algorithmic approaches used (see Figure~\ref{fig:edge_fringe_coherent_blur}).

In our work, to demonstrate our approach, we build a holographic display prototype following the guidance from the recent literature~\cite{shi2021towards}.
Though the general layout of our display is similar to the standard phase-only holographic displays, the more critical details of our implementation are unique to its case as discussed in Section~\ref{sec:proof_of_concept_holographic_display_prototype}.

\subsection{Computer-generated Holography}
CGH deals with computing a hologram that generates the desired light distribution over a target plane when displayed on an SLM \cite{slinger2005computer}.
Hologram calculation with CGH is known to be computationally expensive due to the complexity of physical light simulation models used in CGH~\cite{zhang2020band,zhang2020adaptive}.
The recent advancements in GPUs and deep learning spark the development of new algorithms that promise hologram generation at interactive rates in the future~\cite{shi2021towards}.
Conventional CGH algorithms can be broadly classified using their scene representations.
Such scene representations include point-cloud~\cite{Su:16}, ray~\cite{Wakunami:11}, polygon~\cite{Matsushima:09}, light field~\cite{shi2017near}, and multiplane~\cite{makey2019breaking} representations.
We suggest our readers consult the survey by Corda\etal~\cite{Corda:19} for a complete review of conventional CGH algorithms.

We employ a multiplane representation approach for 3D image reconstructions. 
Our work differs from the rest of the literature~\cite{makey2019breaking,curtis2021dcgh,Mahdi:19,shi2021towards,peng2020neural,choi2021neural3d} in the novel targeting scheme and loss function for the multiplane optimization, which mitigates edge fringe issues and helps reconstruct realistic defocus blur.
Although the work by Choi\etal~\cite{choi2022time} generates realistic defocus using a DMD time-multiplexed SLM, to our knowledge, our work differs as our work targets common liquid-crystal based SLMs. Thus, we believe our work is the first in addressing edge-fringe issues without relying on time-multiplexing.
Our work also differs in calculating phase-only holograms as we show that it can operate with various kinds of CGH methods, including GS~\cite{zhou201530}, SGD~\cite{zhang20173d} and DP~\cite{hsueh1978computer} based approaches.

\begin{figure*}[hbt!]
 \centering
 \includegraphics[width=\linewidth]{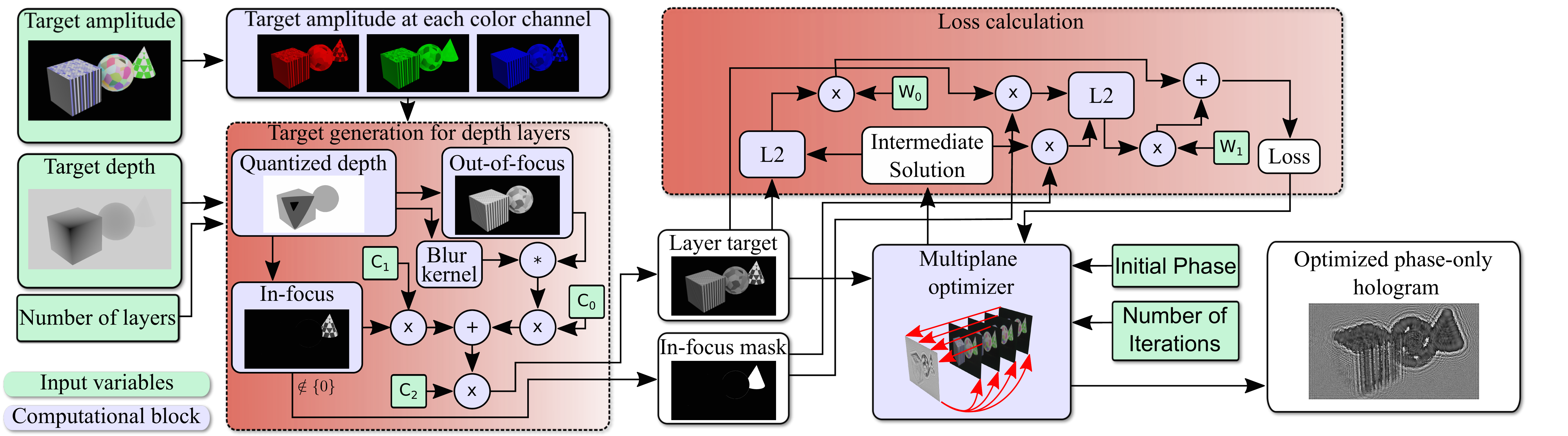}
 \caption{
 Our proposed multiplane computer-generated holography pipeline. 
 We start with an input target amplitude and target depth. 
 Our targeting scheme helps us determine meaningful targets for each depth plane for each color channel. 
 We provide a set of layer targets and masks defining in-focus regions at each layer target to a multiplane optimizer (e.g., Gerchberg-Saxton or Stochastic Gradient Descent). 
 We optimize phase-only holograms using our newly derived loss function.
 }
 \label{fig:method_overview}
\end{figure*}

\section{Targeting Scheme and Loss Function for Multiplane Computer-Generated Holography}
\label{sec:targetting_scheme_loss_function}
We aim to calculate phase-only holograms that simultaneously reconstruct high-quality images at multiple planes at various depth levels.
Here high-quality refers to the generation of multiplane images free from edge fringes with accurate defocus blur.

When a homogeneous, collimated, coherent light source (e.g., a laser) illuminates a phase-only hologram, light diffracts from that phase-only hologram.
Diffracted light interferes and forms the intended images at the target planes in front of the hologram.
To successfully achieve our goals, we must fulfil two primary objectives.
Firstly, we have to identify the target images for each plane at different depth levels. 
Secondly, we must identify a loss function to evaluate our reconstructions at each plane in depth.
\Fig~\ref{fig:method_overview} provides an overview of our targeting scheme and our loss function.

\subsection{Targetting scheme}
A typical multiplane image reconstruction method for CGH requires representing a 3D scene with a target amplitude image, $P_{\mathrm{target}}$ and a target depth, $V_{\mathrm{depth}}$.
In a na\"ive approach~\cite{peng2020neural}, layer targets, $\{P_{\mathrm{layer}_0},...,P_{\mathrm{layer}_n}\}$, are composed of cropped parts of a target amplitude image from each pixel depth level.
Hence, layer targets at each layer, $P_{\mathrm{layer}_k}$, is a sparse image filled with black pixels surrounding in-focus pixels.
In other words, a conventional optimizer for CGH would force the image reconstruction solution to provide black pixels at places where defocus blur of other objects should appear in reality.
Specifically, when the target depth planes are packed closely in terms of distances (a few millimeters), coherent image blur dominates the na\"ive approach's solution, leading to noisy reconstructions suffering from visual quality issues.
The recent and common literature that uses optimization~\cite{wu2021adaptive} and machine learning methods~\cite{lee2020deep} widely adopts the na\"ive approach.
Forcing the image reconstruction to provide black pixels instead of natural defocus at defocused regions causes edge-fringe artifacts to appear in the final image as can be observed in the previous literature~\cite{curtis2021dcgh,Mahdi:19,shi2021towards,peng2020neural,choi2021neural3d}. 
Specifically, edge-fringe artifacts become more pronounced across occlusions in a given scene as these parts are also forced to be painted with black rather than the defocused version of the part of a target scene.
Hence, we argue that multiplane image reconstruction is beyond resolving an optimization or a learning problem, and it has to also deal with the problem formulation.
We argue that tailoring target images can help improve multiplane image reconstructions in CGH.

Inspired by Depth-of-Field (DoF) rendering~\cite{leimkuhler2018laplacian} in conventional 2D computer graphics, we introduce a new targeting scheme for identifying target images at each plane in depth.
Commonly, a $V_{\mathrm{depth}}$ comes with an eight-bit pixel depth.
Firstly, we quantize a given $V_{\mathrm{depth}}$.
In a typical quantization step from our method, we quantize $V_{\mathrm{depth}}$ to a pixel depth between one to four bits (at most 16 consecutive planes).
Our choice in the number of planes emerges from an earlier analysis~\cite{akcsit2019manufacturing}, in which RGBD natural image datasets are analyzed by taking into account the human visual system's DoF.
Quantized depth images help us identify regions from $P_{\mathrm{target}_k}$ as in-focus regions, $P_{\mathrm{focus}_k}$, and out of focus, $P_{\mathrm{out}_k}$, at each depth level, $k$.
We convolve regions of $P_{\mathrm{out}_k}$ with a Gaussian kernel to achieve a defocus similar to an incoherent case, $P_{\mathrm{defocus}_k}$.
Here, the Gaussian kernel's size is in inverse relation with the distance to the focus plane.
We sum $P_{\mathrm{defocus}_k}$ and $P_{\mathrm{focus}_k}$ using weights, leading to the final target form,
\begin{equation}
P_{k}=w_2 (w_0 P_{\mathrm{defocus}_k} + w_1 P_{\mathrm{focus}_k}),
\end{equation}
where $P_{k}$ is the target image at k-th plane, and $w_0, w_1, w_2$ represents weights.
These weights help us control brightness levels of sharp parts, blurry parts and overall image.
To our knowledge, in multiplane CGH, our targeting scheme is the first to add defocus-blur to target images.

\subsection{Loss Function}
Evaluating multiplane image reconstructions requires a loss function.
Evaluating the image quality of the reconstructed images often involves measuring L2 distance between each element of layer targets, $P_{\mathrm{plane}_k}$ and reconstructed images $I_{\mathrm{plane}_k}$.
We argue that a multiplane image reconstruction problem demands a more sophisticated image quality metric.
Our argument originates from the fact that different parts of the image will come into focus at each depth level while the remaining parts have to be blurry.
Hence, we argue that in-focus regions require more attention in precision than out-of-focus parts.
Thus, we propose a new loss function, $\mathcal{L}_m$, a weighted sum of two different loss functions,
\begin{equation}
\begin{split}
\mathcal{L}_{m}=m_0 \mathcal{L}_2(P_{k},I_{k}) & +m_1 \mathcal{L}_2(M_{k} \odot P_{k}, M_{k} \odot I_{k}),
\end{split}
\end{equation}
where $m_0,m_1$ represents weights,  $I_k$ represents a reconstructed image at a k\textsuperscript{th}  plane, and $M_k$ represents a binary mask highlighting only the sharp parts in a k\textsuperscript{th}  plane.
We survey various values of $m_0$ and $m_1$ in a brute-force fashion.
In our practical observations, choosing $m_0=1.0, m_1=2.1$ leads to the best looking visuals in our holographic display.
Weighted with $m_0$, $\mathcal{L}_2(P_{k},I_{k})$ part of $\mathcal{L}_{m}$ is a standard L2 norm that evaluates the entire image with respect to a target image.
The $m_1 \mathcal{L}_2(M_{k} \odot P_{k}, M_{k} \odot I_{k})$ part of $\mathcal{L}_{m}$ evaluates the in-focus part of an image with respect to a target image (L2-focus).
For accounting gaze-contingency in this loss, see our supplementary materials.

\section{Calculating Multiplane Holograms}
\label{sec:calculating_multiplane_holograms}
The required ingredients for calculating phase-only holograms are all introduced at this point, including our targeting scheme, our loss function and a model for propagating light (see our supplementary materials).
The routines discussed here are valid for a phase-only hologram, $O_{h}$, illuminated by a coherent beam, $U_{i}$.

We will deal with reconstructing 3D images in the whereabouts of an SLM, leading to light propagation distances, $r$, from $0~mm$ to a few millimetres.
Commonly, in this regime, people use DP~\cite{hsueh1978computer}
method to encode a complex field into a phase-only hologram (see Maimone\etal~\cite{maimone2017holographic} for a simplified formulation).
Across the literature~\cite{maimone2017holographic,shi2021towards}, images reconstructed at this regime provide the best known visual quality.
We can use our targeting scheme with the Double Phase method.
In that case, we couple each layer target, $P_{\mathrm{layer}_k}$, with a random phase to generate a target field. 
Then, we propagate each target field from its plane ($r\approx30~cm$ in our simulations) to a phase-only hologram plane.
We sum up all the propagated fields from each target at the hologram plane.
Finally, we follow the same routine from recent works~\cite{maimone2017holographic,shi2021towards} by shifting the field towards SLM plane ($r\approx-30~cm$) and applying DP encoding.
However, combining the DP with our targeting scheme would not entirely resolve the raised issues in holographic displays.
We provide actual evidence in our evaluation section accordingly.

\begin{figure*}[hbt!]
\centering
\includegraphics[width=\linewidth]{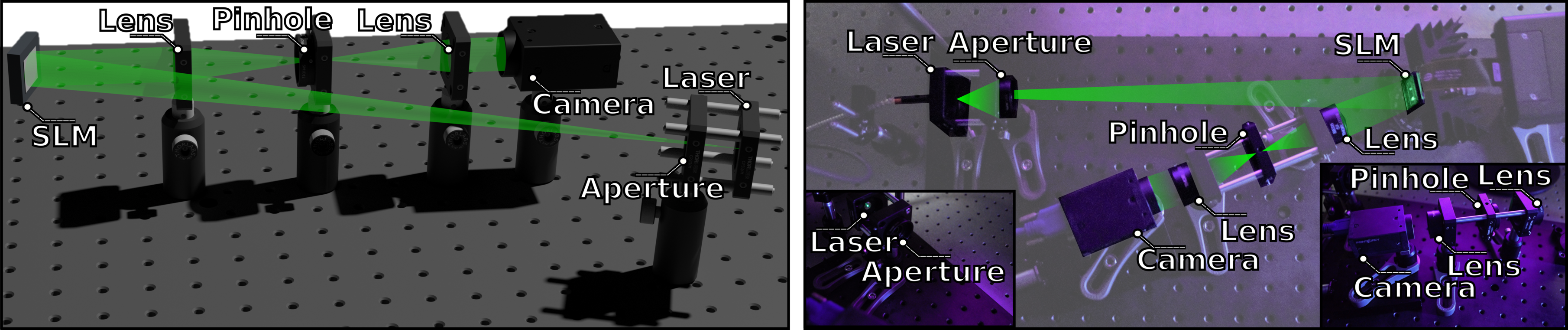}
\caption{
Our proof-of-concept holographic display.
(Left) A 3D layout of our design showing the arrangement of the optical and optomechanical components of our prototype.
(Right) A photograph of our proof-of-concept holographic display setup with annotations of primary components.
Both at the layout and photographs, light direction and path are indicated with a green line.
The 3D layout model of our prototype is made available at our codebase~\cite{realistic_defocus_repo}. 
}
\label{fig:hardware_setup}
\end{figure*}

Alternatively, in this regime, a typical optimization method would have to first optimize a hologram for some propagation distance, then rely on shifting and DP coding.
We ask ourselves if a hologram could be directly optimized on the SLM plane.
Given that the propagation distance is small, the conventional beam propagation methods~\cite{heurtley1973scalar} would not hold well or provide a meaningful result at near zero distance.
We find out that if we propagate a phase-only hologram from an SLM plane some distance (e.g., $r=30~cm$) and propagate it back to the hologram plane (e.g., $r=-30~cm$), we can reconstruct an image on an SLM plane, leading to a formulation,
\begin{equation}
u(x,y) = ( O_{h}(x,y,) * h_{r}(x,y) ) * h_{-r}(x,y),
\label{equ:forward}
\end{equation}
where $h_{r}$ and $h_{-r}$ represent light transport and $u$ represents the reconstructed image.
The composition of these propagations would be the identity since propagation is inverted by backpropagation. 
However, in this case, the wavefront at the SLM can be optimized because the wavefront at a distance r must be cropped to a region that does not include the full wavefront of that plane. 
Physically, this would correspond to having an aperture at a distance after the SLM.
Using this forward model, we optimize a phase-only hologram using SGD implementations from recent literature~\cite{Chakravarthula:20,peng2020neural}.
Our findings suggest that this approach would lead to noisy image reconstructions.
We fix this noise issue by constraining the phase updates of SGD, $\phi$ of $O_{h}$, with an approach inspired from the DP method.
Our phase values, $\phi$, follows,
\begin{equation}
\begin{split}
\phi_0 = \phi - \bar{\phi} \\
\phi_\mathrm{low} = \phi_0 - \mathrm{offset} \\
\phi_\mathrm{high} = \phi_0 + \mathrm{offset} \\
x_\mathrm{even}, y_\mathrm{even} \in \{ 0, 2, 4, 6, \dots \} \\
x_\mathrm{odd}, y_\mathrm{odd} \in \{ 1, 3, 5, 7, \dots \} \\
\phi[x_\mathrm{even}, y_\mathrm{even}] = \phi_\mathrm{low}[x_\mathrm{even}, y_\mathrm{even}] \\
\phi[x_\mathrm{odd}, y_\mathrm{odd}] = \phi_\mathrm{low}[x_\mathrm{odd}, y_\mathrm{odd}] \\
\phi[x_\mathrm{even}, y_\mathrm{odd}] = \phi_\mathrm{high}[x_\mathrm{even}, y_\mathrm{odd}] \\
\phi[x_\mathrm{odd}, y_\mathrm{even}] = \phi_\mathrm{high}[x_\mathrm{odd}, y_\mathrm{even}] \\
O_{h} \leftarrow \phi,
\end{split}
\label{equ:phase_update_rule}
\end{equation}
where offset is a variable to be optimized.

Since there is a readily available differentiable version of beam propagation~\cite{aksit_kaan_2022_6528486}
compatible with widely used machine learning libraries~\cite{paszke2017automatic}, optimizing phase-only holograms becomes even an easier task.
We provide a pseudo-code for our SGD optimization routine as in Listings 1.

\begin{figure}[t]
\centering
\begin{minipage}{.44\textwidth}
\begin{lstlisting}[language=iPython,escapeinside={(*}{*)},caption={Stochastic-Gradient based multiplane phase-only hologram optimization algorithm when reconstructing images at a spatial light modulator plane. The abstraction is Pythonic. Note that this optimization runs for each color channel separately.},label=list:sgd_multiplane]
import torch.optim as optim
from RemovedForAnonymity import propagate_beam,generate_complex_field

# Provide an initial phase for a hologram (random, manual or learned).
(*\textbf{$\phi$}*) = define_initial_phase(type='random') 
(*\textbf{$\phi$}*).requires_grad = True
# Provide number of iterations requested.
iter_no= 60
# Setup a solver with 
optimizer = optim.Adam([{'params': (*\textbf{$\phi$, $offset$}*)}], lr=0.04)
# Calculate targets for each plane.
(*\textbf{$P_0,P_1,P_2,...,P_n$}*) = targetting_scheme(distances)

# Iterates until iteration number is met.
for i in range(iter_no):
    # Distances between a hologram and target image planes.
    for distance_id,distance in enumerate(distances):
        # Clearing gradients.
        optimizer.zero_grad()
        # Phase constrain (Equation 5).
        (*\textbf{$\phi$}*) = phase_constrain((*\textbf{$\phi$}*), (*\textbf{$offset$}*))
        # Generates a hologram with the latest phase pattern.
        (*\textbf{$O_h$}*) = generate_complex_field(1., (*\textbf{$\phi$}*))
        # Forward model (e.g. distance=30 cm, delta=1 mm).
        (*\textbf{$K$}*) = propagate((*\textbf{$O_h$}*), distance)
        (*\textbf{$U$}*) = propagate((*\textbf{$K$}*), -distance + delta)        
        # Calculating loss function for the reconstruction.
        loss += (*\textbf{$\mathcal{L}_{m}$}*)((*\textbf{$|U|^2$}*), (*\textbf{$P_{(distance\_id)}$}*))
        # Updating the phase pattern using accumulated losses.
        loss.backward()
        optimizer.step()
    
# Optimized multiplane hologram:
(*\textbf{$\phi \rightarrow O_h$}*)
\end{lstlisting}
\end{minipage}
\label{list:sgd_multiplane}
\end{figure}

\section{Holographic Display Prototype}
\label{sec:proof_of_concept_holographic_display_prototype}
To demonstrate that we can leverage our CGH approaches in practice, we build a proof-of-concept holographic display prototype using off-the-shelf components.
We provide a detailed overview of our optical schematic and photographs from our experiment bench in \Fig~\ref{fig:hardware_setup}.

\begin{figure*}[hbt!]
\centering
\includegraphics[width=\linewidth]{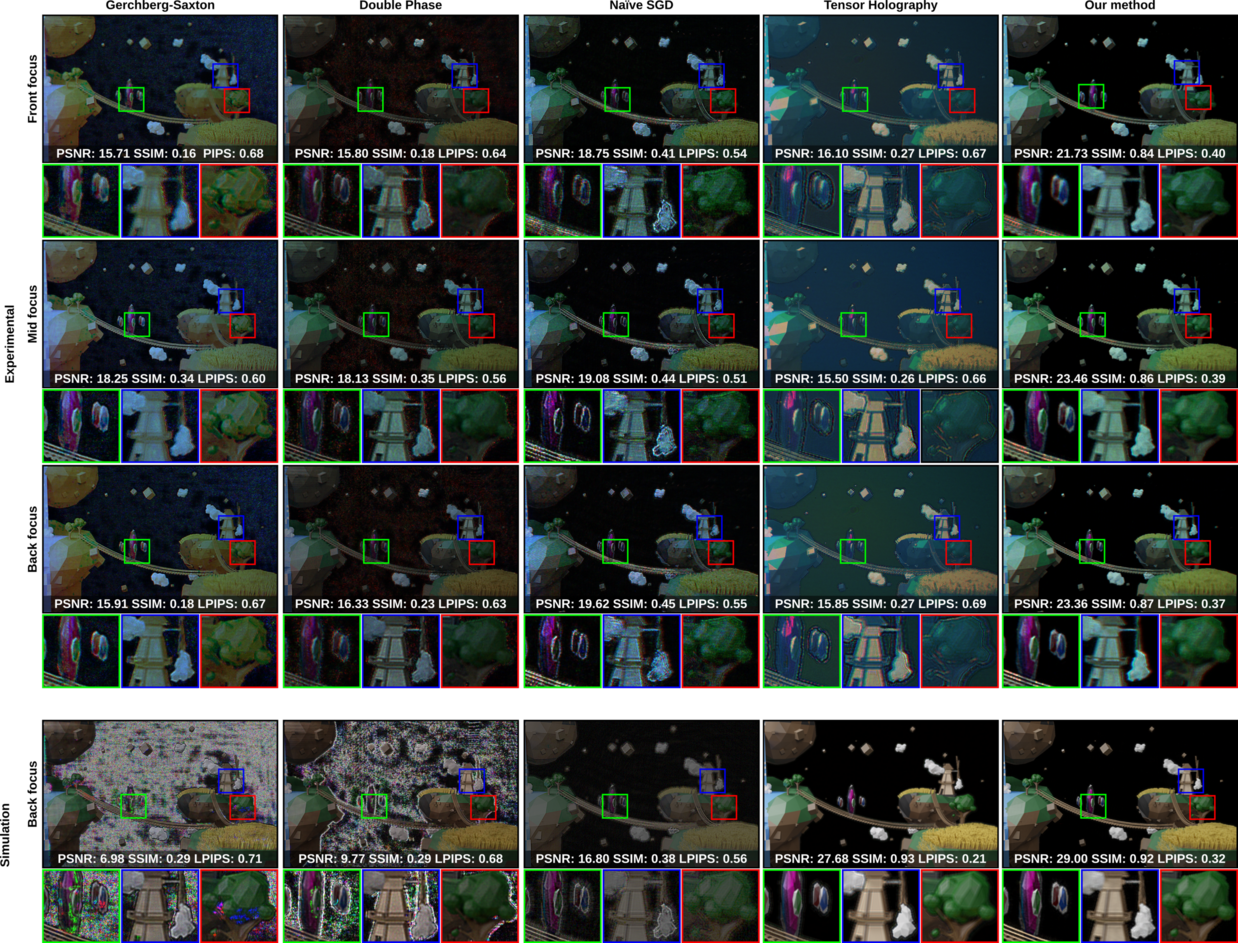}
\caption{
Comparison of methods.
We show actual captures from our holographic display for three different focuses in the top three rows. 
In addition, a result from a simulation focusing on the back is presented in the bottom row.
All the captures from our holographic display use 20 ms exposure time.
We tune laser powers for each case highlighted per column to get to the best possible image quality in the corresponding case.
These cases include the following items from left to right: (1) optimizing multiplane holograms using the Gerchberg-Saxton method~\cite{zhou201530} with our targeting scheme, (2) optimizing multiplane holograms using the Double Phase method~\cite{hsueh1978computer}
with our targeting scheme, (3) optimizing multiplane holograms using Stochastic-Gradient Descent (SGD)~\cite{zhang20173d} with na\"ve targeting where defocus parts are painted with black pixels in the target images.
This specific case approximates the na\"ve version of the work by Peng\etal~\cite{peng2020neural}, Chakravarthula\etal~\cite{Chakravarthula:20} and Choi\etal~\cite{choi2021neural3d} closely, (4) Optimizing a continuous depth hologram using the state-of-the-art method, tensor holography~\cite{shi2021towards}, and (5) optimizing a multiplane hologram using our method uses SGD and use an update rule inspired from Double Phase method.
In actual captured results and simulations, our method distinguishes itself as the contrast preserving, fringe artifacts free and high-visual quality method.
}
\label{fig:comparison}
\end{figure*}

\paragraph{Optical and Optomechanical Assembly.}
The optical path of our prototype starts from a single-mode fiber-coupled multi-wavelength laser light source, LASOS MCS4, which combines three separate laser light sources peeking at 473~nm, 515~nm and 639~nm.
We limit the diverging beams coming out of our fiber with a pinhole aperture, Thorlabs SM1D12. 
After this pinhole aperture, light beams reach our phase-only SLM, Holoeye Pluto-VIS.
The phase-modulated beam arrives at a 4f imaging system composed of two 50~mm focal length achromatic doublet lenses, Thorlabs AC254-050-A, and a pinhole aperture, Thorlabs SM1D12, removing undiffracted light. 
We capture the image reconstructions with an image sensor, Point Grey GS3-U3-23S6M-C USB 3.0, located on an X-stage (Thorlabs PT1/M travel range: 0-25 mm, precision: 0.01 mm).

Our holographic display prototype is configured as an off-axis imaging system. 
We are using the half diffraction order location for our optical reconstructions. 
Our desired half diffraction order beam is on-axis with respect to the 4f imaging system, and the illumination beam is slightly off-axis. 
We rely on a linear grating term which will be explained in the \textit{Computation and Control Modules} paragraph. 
Therefore, a beamsplitter in front of SLM and a linear polariser are not required for our prototype.

\begin{figure*}[hbt!]
\centering
\includegraphics[width=\linewidth]{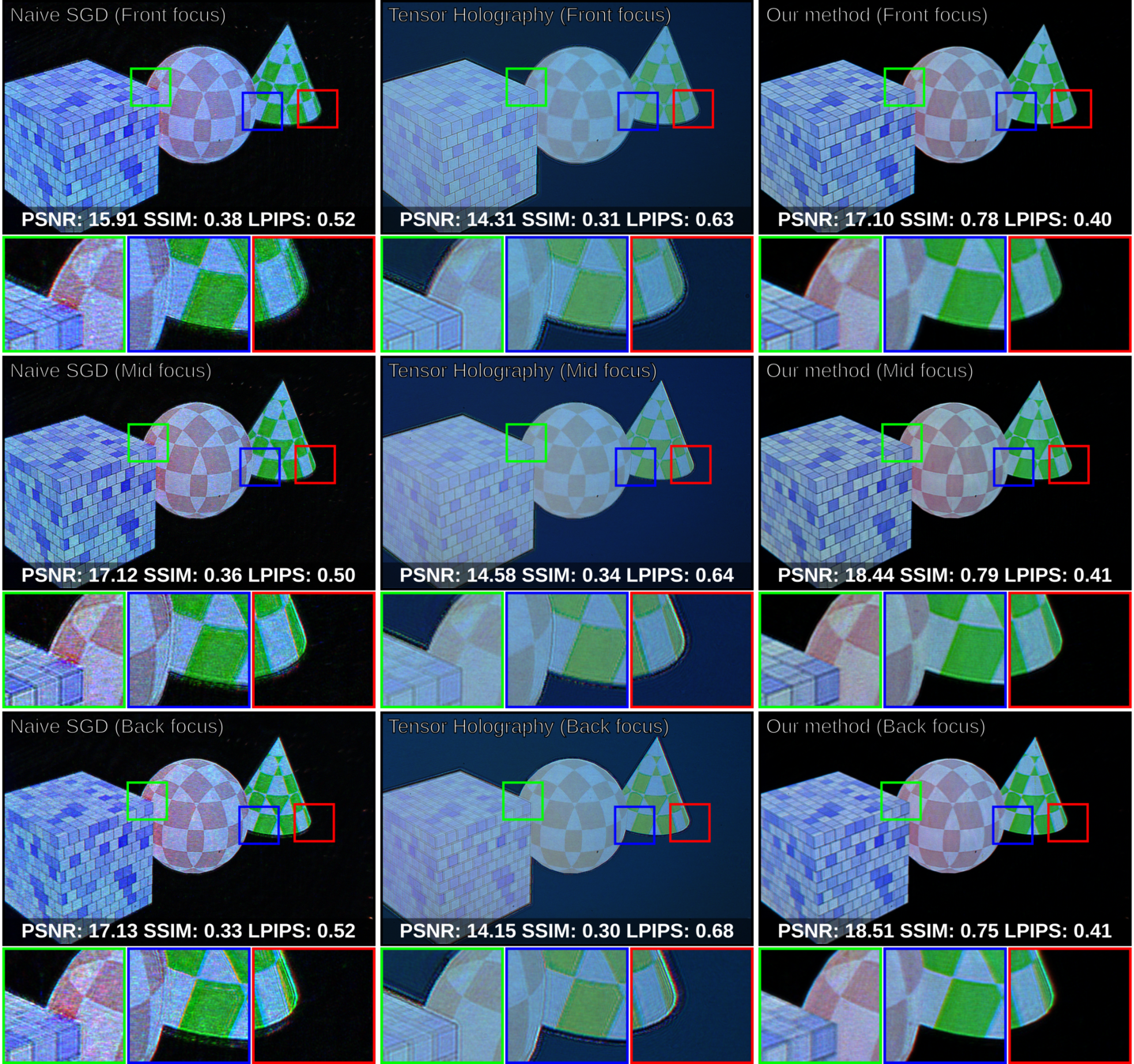}
\caption{
Fringe free Computer-Generated Holography in edges and edge cases.
Representing 3D scenes in holographic displays often yields artifacts around the corners of defocused objects. 
Specifically, such artifacts are more pronounced in scenes populated with black content.
Here, we show actual captures from our holographic displays with 20 ms exposure and the same laser power settings.
These captures show how these artifacts would be an issue with the most current state of the art methods in the literature for various focus states presented in each row.
The first two columns include a Stochastic-Gradient Descent based approach (e.g., \cite{peng2020neural}) with na\"ve targetting and tensor holography~\cite{shi2021towards}.
The third column shows capture for our result, where no such artifact is visible, and transition between planes approximates incoherent case better.
}
\label{fig:sparse_content}
\end{figure*}

\paragraph{Computation and Control Modules.} 
We use a computer with NVIDIA GeForce RTX 3070 laptop GPU with 8~GB memory and an Intel i7, 4.6 GHz CPU to drive our holographic display prototype.
To avoid undiffracted light, we update calculated $O_h$ with a linear phase grating term,
\begin{equation}
  O_h'(x,y) =
  \begin{cases}
            e^{-j(\phi(x,y) + \pi)} & \text{if $y=$ odd} \\
            e^{-j\phi(x,y)}         & \text{if $y=$ even} \\
  \end{cases}
\end{equation}
where $\phi$, $x$, $y$ represents the original phase of $O_h$.

\section{Evaluation}
\label{sec:evaluation}
In all our experiments, we use the exact exposure times, $20~ms$, and distances between our target planes set to $1~mm$.
For example, if we use four planes, the reconstructed images will correspond to a volume with $4~mm$ ($\pm 2~mm~\mathrm{to~SLM}$). 
For more results, please to consult our supplementary materials.

\paragraph{Defocus blur}
A sample experimental and simulation comparison is provided as in Figure~\ref{fig:comparison}.
The na\"ive multiplane targeting approach, where defocus parts are painted with black pixels in the target image, forms a baseline for our comparison.
When GS and DP are used with n\"aive multiplane targeting, they lead to poor image quality.
However, na\"ive multiplane targeting with SGD~\cite{Chakravarthula:20,peng2020neural} provided a result with more reasonable image quality. 
Thus, we discard GS and DP with na\"ive targeting from our discussion.
\textit{Instead, we use GS and DP methods with our targeting schemes.}
Although these options with our targeting lead to a more reasonable visual quality, it is still lacking compared with state of the art.
Finally, we also add the tensor holography~\cite{shi2021towards} to our comparison.
Our experimental results showed that our method using SGD optimizations with a DP inspired phase update rule provided the most appealing image quality with realistic looking defocus blur.
Our experimental findings are in line with our simulation results in our assessment.
The simulation case for tensor holography has the contour outlines in defocus regions but not pronounced as in the experimental result from our evaluation and the original work.

\paragraph{Edge fringe}
The edge fringe issue in CGH is typically highly apparent in contents dominated by black pixels, making such content an excellent way to quantify our improvement over existing literature.
In Figure~\ref{fig:sparse_content}, we compare our method against the state of the art while targeting sparse content dominated by black pixels.
Our method distinguishes itself as an edge fringe artifact free result.

\paragraph{Blur size and number of depth planes.}
Our multiplane CGH generation pipeline can generate images with various multiple quantization levels in depth.
In addition, our CGH pipeline offers control over blur size, potentially leading to rendering scenes in a styled way according to a viewer's taste.
In our method, Gaussian Kernels can be replaced with Zernike polynomials in the future to support the prescription of a viewer.
We provide the evidence that our  CGH pipeline can provide images with various quantizations and blur sizes as in our supplementary materials.

\paragraph{Required computational resources.}
For our method, optimizing a full color hologram for a scene with eight target planes takes 57 seconds with a memory footprint of 4623 MB using NVIDIA GeForce RTX 3070 laptop GPU with 8~GB memory.
In comparison, our baseline with naive targeting takes 55 seconds to run while requiring 4587 MB on the same GPU.
It should be noted that tensor-holography~\cite{shi2021towards} is significantly faster than optimization methods as it replaces the iterative optimizations with a network.
For our case it takes almost 1.5 second to generate a full color hologram and requires 5908 MB of memory.

\paragraph{Projection distances.}
The standard literature in the previous ten years mainly relied on generating images away from an SLM, either inside or in front of it (e.g.,~\cite{peng2020neural,kim2022holographicglasses}).
Specifically, augmented reality near-eye display designs that place SLMs in front of an eye~\cite{kim2022holographicglasses} could benefit from improved CGH methods that work better in long projection distance cases as these displays generate images inside an SLM.
Typically, the edge fringe artifacts become more apparent, and image quality gets worse when images are reconstructed away from an SLM.
These issues make the long projection distance case a perfect base for analyzing the quality of CGH methods.
In our pipeline, if we update our forward model in Equation~\ref{equ:forward} by dropping the back projection part, leading to
$u(x,y) = O_{h}(x,y,) * h_{r}(x,y)$,
and if we drop our phase update rule in Equation~\ref{equ:phase_update_rule}, we can use our multiplane CGH pipeline for long projection distances.
In our supplementary, we provide an evidence that our method improves image quality also in long projection distances.
We provide \textit{an evidence for compatibility to AR near-eye display}, our readers can consult to our supplementary material for this evidence.

Using our proof-of-concept holographic display, we demonstrated a complete multiplane CGH calculation pipeline with near-accurate defocus blur.
While we noticeably mitigate edge fringe issues and our results approximate natural focus blur, there are remaining issues and potential new research paths.
These issues and research paths includes occlusion in multiplanar holography, perceptual holography and practical issues.
We expand on these topics with a discussion included in our summplementary.

\section{Discussion}
\label{sec:discussion}
We demonstrated a new CGH method that introduces a novel loss function, targeting scheme, and optimization method for multiplane image reconstructions.
In addition, we also built a proof-of-concept holographic display to assess the experimental performance of our introduced algorithmic methods.
While we noticeably mitigate edge fringe effects and our results approximate natural focus blur, there are remaining issues and potential new research paths that can help us converge to better implementations in the future.

\paragraph{Perceptual Holography.}
Recent literature \cite{9756777,chakravarthula2021gaze,cem2020foveated} promises to take advantage from the qualities of human visual system by introducing new loss functions that help reconstruct images at peripheral vision with perceptual guidance and gaze-contingency.
Although these works open up a new combination in research, these works can be best described as first stabs at achieving true perceptual realism.
Future perceptual graphics and computational displays research must find ways to carefully tune the image spatiotemporal qualities of holography to achieve life-like visuals.

\paragraph{Occlusion in multiplanar holography.}
A change in a user's point-of-view when observing multiplane images can lead to occlusion related issues with most multiplane CGH approaches, including ours.
These occlusions manifest as missing image parts or images that look ``holo'' (seeing one object inside another).
Hence, addressing the occlusion issue is essential and an outstanding scientific question in multiplane CGH.
Further research in data representations in multiplane is also required to provide occlusions similar to point-based representations in the literature~\cite{shi2021towards}.

\paragraph{Visual quality metrics.}
We strongly argue that commonly used image quality assessment metrics such as PSNR and SSIM do not apply to multiplane holograms for various obvious reasons.
For instance, there is no metric to assess the defocus blur in 3D scenes. 
The values coming out of such metrics do not entirely correlate with the experience of a human subject (e.g., marking blurry as better to sharper images with speckle noise).
The community can benefit from a new set of metrics tailored for 3D CGH.

\paragraph{Learned Holography.}
Learning methods have garnered interest among the CGH community.
An exciting line of research from work by Lee\etal~\cite{liu2021deep} promises to benefit Variational Autoencoders (VAE) for generating complex holograms.
Though their work targets lower resolution image reconstructions on a single plane, VAEs may promise a dimensionality reduction in our problem (smaller network).
On the other hand, there is active research on replacing optimization steps and further enhancing optimization routines with learned-references~\cite{Hyder:20}.
Optimizer based solutions (e.g., SGD) can provide much higher quality holograms concerning end-to-end hologram generation networks.
Accelerating the optimizations with unrolled networks rather than deriving an end-to-end solution for hologram generation can also provide a gateway towards higher quality image reconstructions that also arrive with the benefit of speed.
For us, improving optimization routines with networks stands out as an exciting opportunity.

\paragraph{Practical issues}
Despite the well-known advantages of CGHs, there are still many challenges to be addressed, including field-of-view~\cite{kuo2020high}, depth-of-field~\cite{makey2019breaking}, and speckle noise~\cite{slinger2005computer}. 
Another challenge in working with a laser-based display system is dust and imperfections on optical components such as diamond turning marks, which causes diffraction problems and undesired fringe patterns on the holograms.
We believe holographic displays need to support imperfect optical components to be practical and highly reproducible at the consumer level.
These practical issues may have seen as engineering hurdles, but we believe these issues fall into the category of tolerance related research in optical and algorithmic designs.

\section{Conclusion}
\label{sec:conclusion}
The display and graphics technologies industry is, in many ways, well established, yet the future remains unsettled.
Innovations leading to seamless blends of graphically created digital 3D objects with the physical natural world will disrupt the status quo.

To fulfil the demands of this ultimate goal, we evaluate CGH as the future display and graphics technology.
Specifically, we study representing 3D scenes as multiplane images using CGH without the inherent coherent artifacts such as fringes or incorrect defocus blur.
Our approach provides a phase-only hologram that can simultaneously reconstruct images at various depths with the near-correct optical focus cues.
While we offer a unique solution to some of the fundamental challenges of coherent display systems, hurdles remain in CGH implementations to claim superiority over other display and graphics technologies. 
We hope to inspire the relevant research communities to investigate CGH as the ultimate display and graphics method.

\section*{Supplementary material}
The code base of our framework is available at our repository~\cite{realistic_defocus_repo}.
We provide the source files of our figures in our supplementary documentation.

\section*{Acknowledgement}
The authors would like to thank reviewers for their valuable feedback.
We would like to thank Erdem Ulusoy and G\"{u}ne\c{s} Ayd{\i}ndo\u{g}an discussions in the early phases of the project; 
Tim Weyrich and Makoto Yamada for dedicating GPU resources in various experimentation phases;
David Walton for their feedback on the manuscript;
Yuta Itoh is supported by the JST FOREST Program Grant Number JPMJPR17J2 and JSPS KAKENHI Grant Number JP20H05958 and JP21K19788.
Hakan Urey is supported by the European Innovation Council's HORIZON-EIC-2021-TRANSITION-CHALLENGES program Grant Number 101057672 and Tübitak’s 2247-A National Lead Researchers Program, Project Number 120C145..
Kaan Ak\c{s}it is supported by the Royal Society's RGS\textbackslash R2\textbackslash 212229 - Research Grants 2021 Round 2 in building the hardware prototype.

\bibliographystyle{abbrv-doi}

\bibliography{references}

\NewEmptyPage


\section*{Supplementary material}

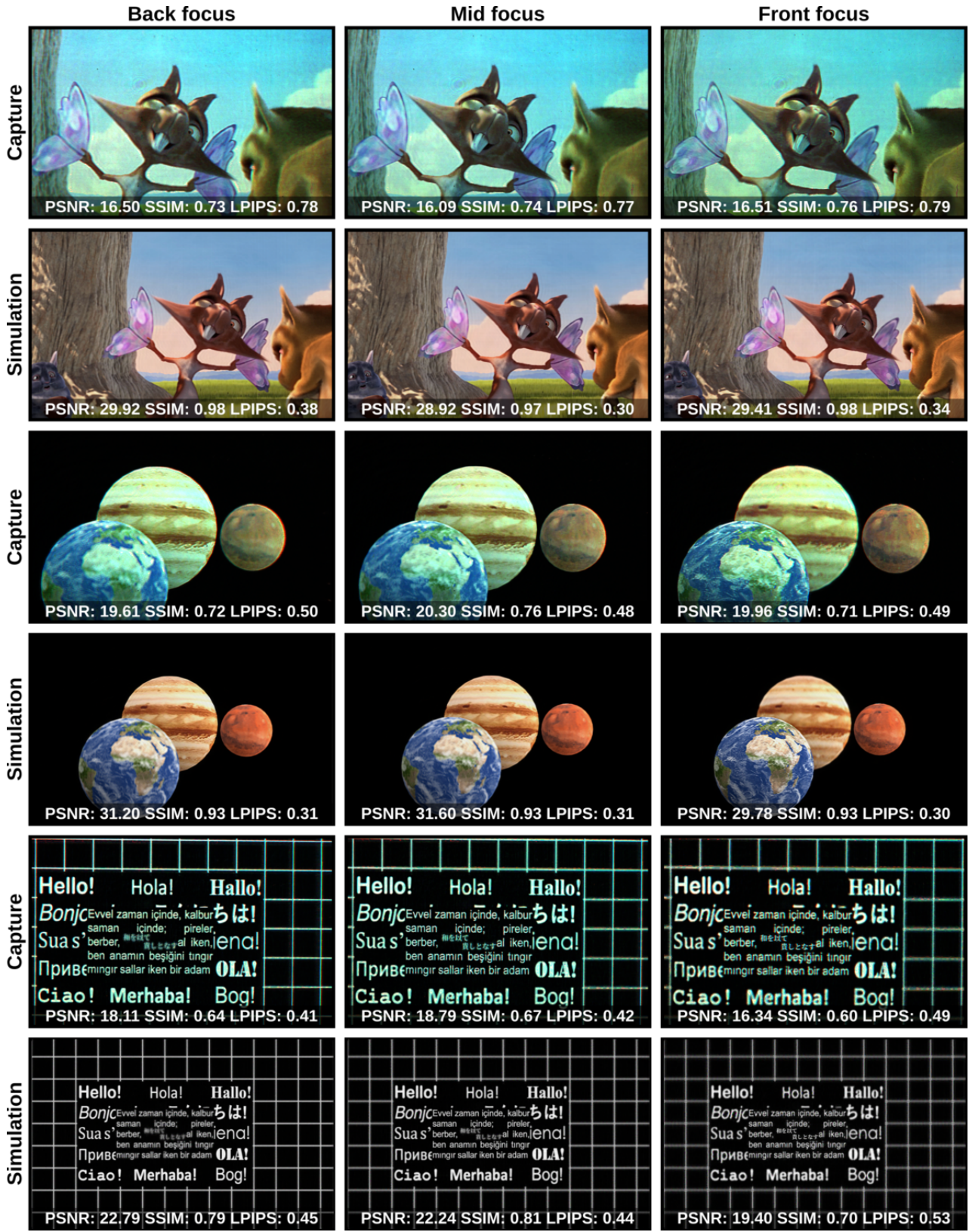

Figure 3: Additional capture and simulation results of our method with various different scenes and focuses. The provided captures are generated on the Spatial Light Modulator (SLM) plane. All the images are captured with 20 ms exposure time using our holographic display. The provided simulated images are also generated on the Spatial Light Modulator (SLM) plane. All the simulated images are results from our hologram reconstruction simulation using our method. Both in captures and simulations, there are six target image planes in depth with 1 mm separation between each depth plane (-2.5 mm and +2.5 mm with respect to an SLM).

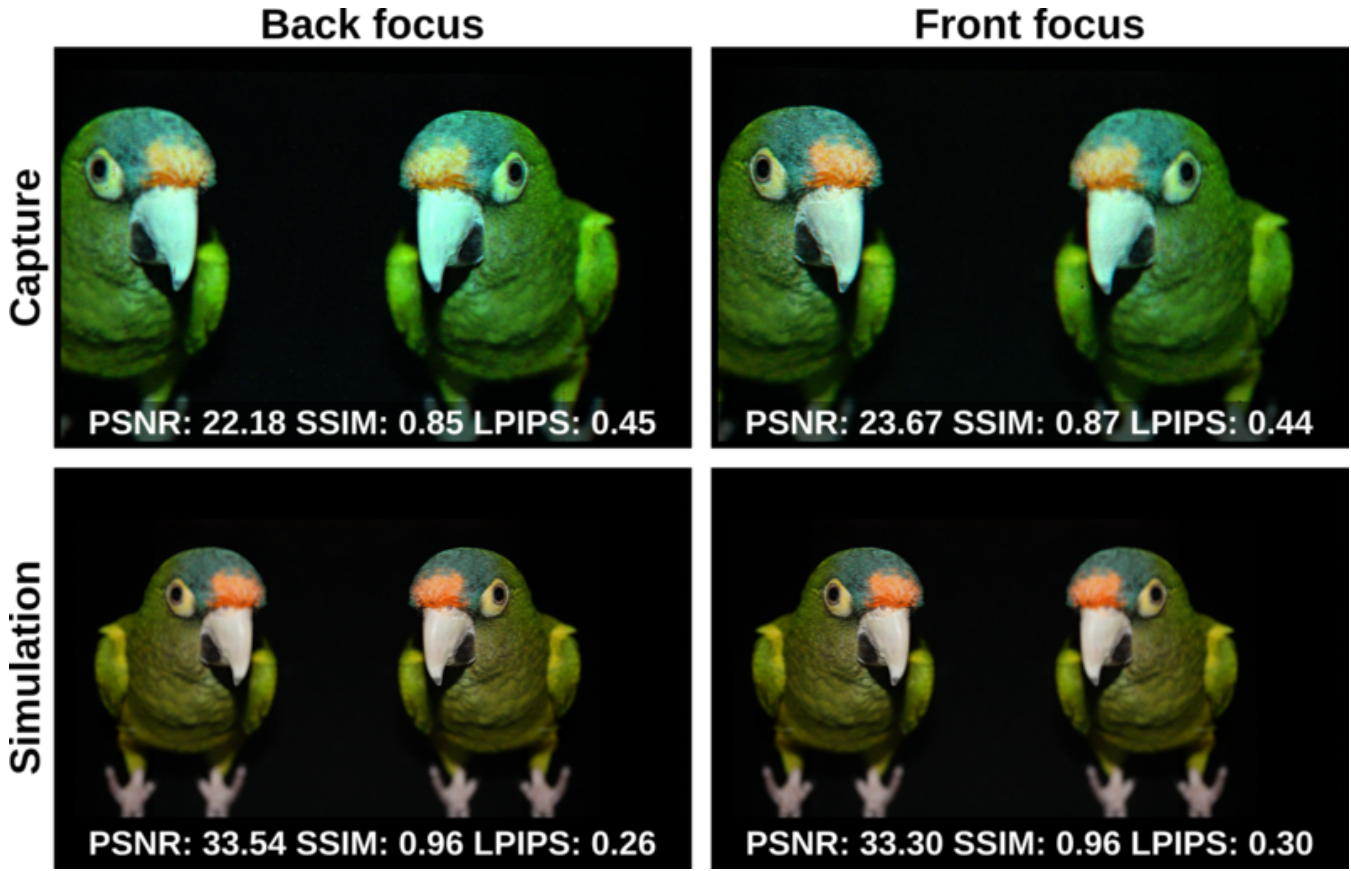

Figure 4: Additional capture and simulation results of our method with various different scenes and focuses. The provided captures are generated on the Spatial Light Modulator (SLM) plane. All the images are captured with 20 ms exposure time using our holographic display. The provided simulated images are also generated on the Spatial Light Modulator (SLM) plane. All the simulated images are results from our hologram reconstruction simulation using our method. Both in captures and simulations, there are six target image planes in depth with 1 mm separation between each depth plane (-2.5 mm and +2.5 mm with respect to an SLM).

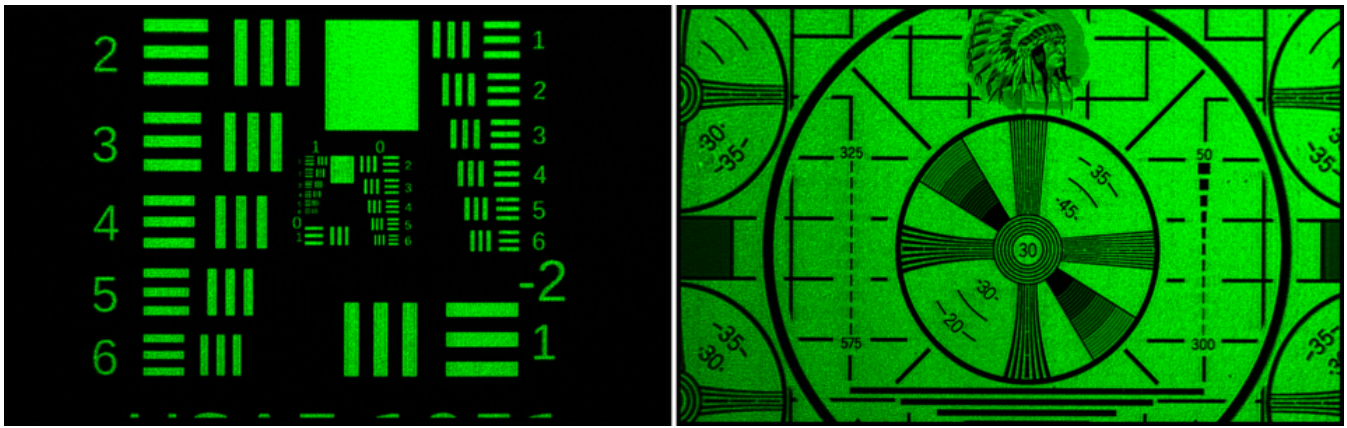

Figure 5: Additional 2D captures. USAF 1951 and Indian Head Pattern charts demonstrate our holographic display prototype's resolution and contrast capability.

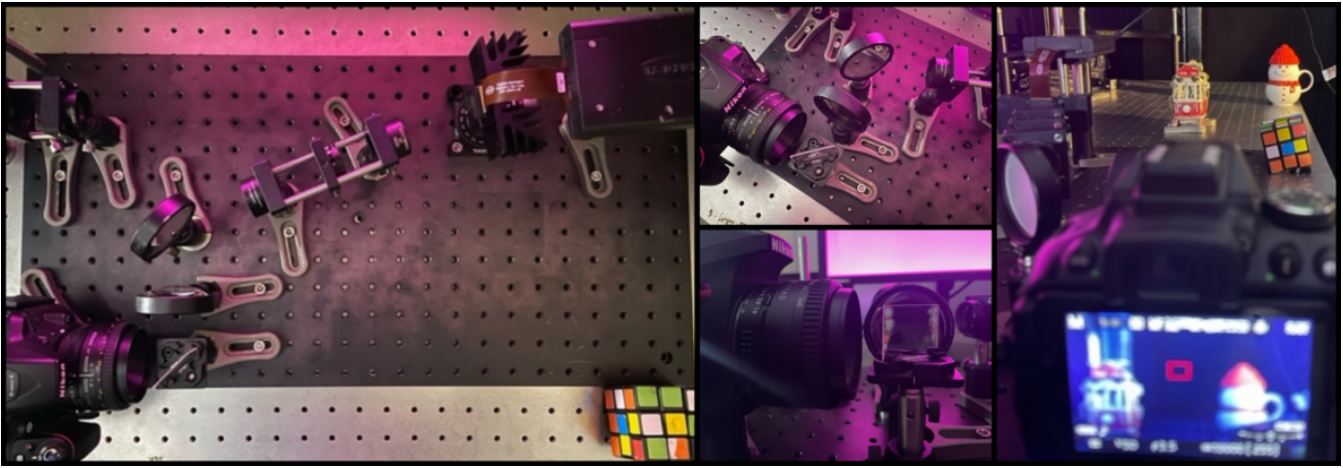

Figure 6: Our proof-of-concept holographic display for Augmented Reality Near-Eye Display application example. Here, we provide various captures of our modified holographic display for AR near-eye display application. We additionally introduce a mirror, eyepiece lens and beam splitter to our display. The camera and the lens imitate the user's view with changing focus cues. The right image demonstrates the actual scene that we used in our captures. The Rubik's Cube, train, and snowman are located 45 cm, 65 cm, and 100 cm away from the camera.

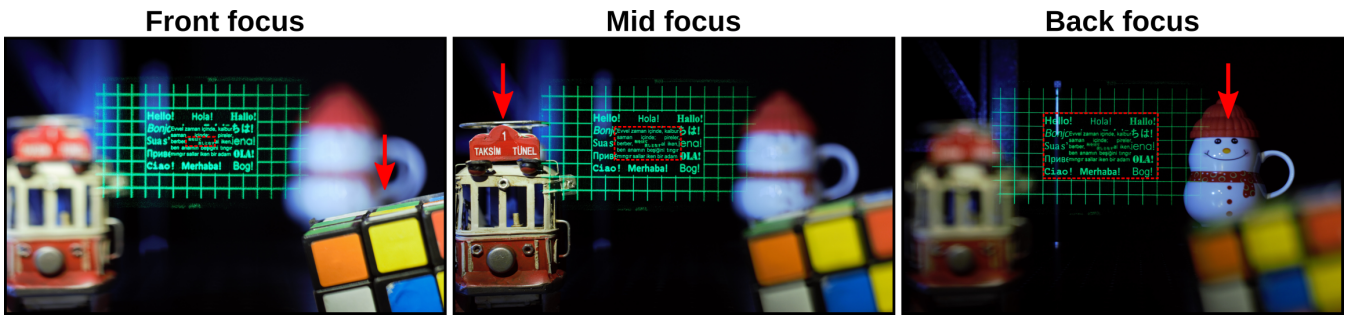

Figure 7: Augmented Reality Near-Eye Display application example. Here, we provide samples from our modified holographic display approximating a baseline AR near-eye display. We show virtual images generated at various optical depths in the given scene. There are eight target image planes in depth with 1 mm separation between each depth plane (-3.5 mm and +3.5 mm with respect to an SLM). The focused virtual image planes correspond to 45 cm, 65 cm, and 100 cm with respect to the user's view. We indicate these plane locations with real objects (Rubik's Cube, train, and snowman) that are placed on the target distances. The red boxes and arrows indicate the focused depth plane in the captures.

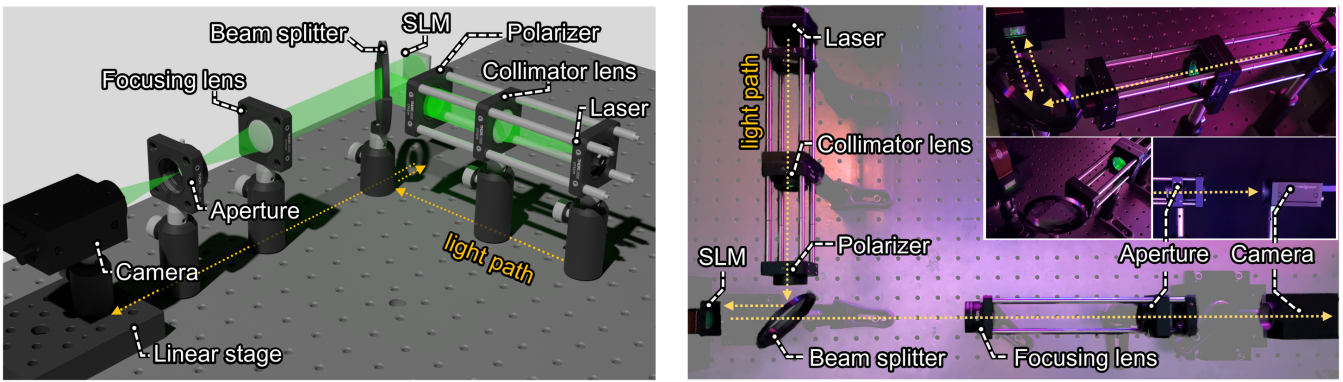

Figure 8: Our proof-of-concept holographic display for long projection distances. (Left) A three-dimensional layout of our design shows the arrangement of our prototype's optical and optomechanical components. (Right) A photograph of our proof-of-concept holographic display setup with annotations of primary components. Both at the layout and photographs, light direction and path are indicated with a yellow line.

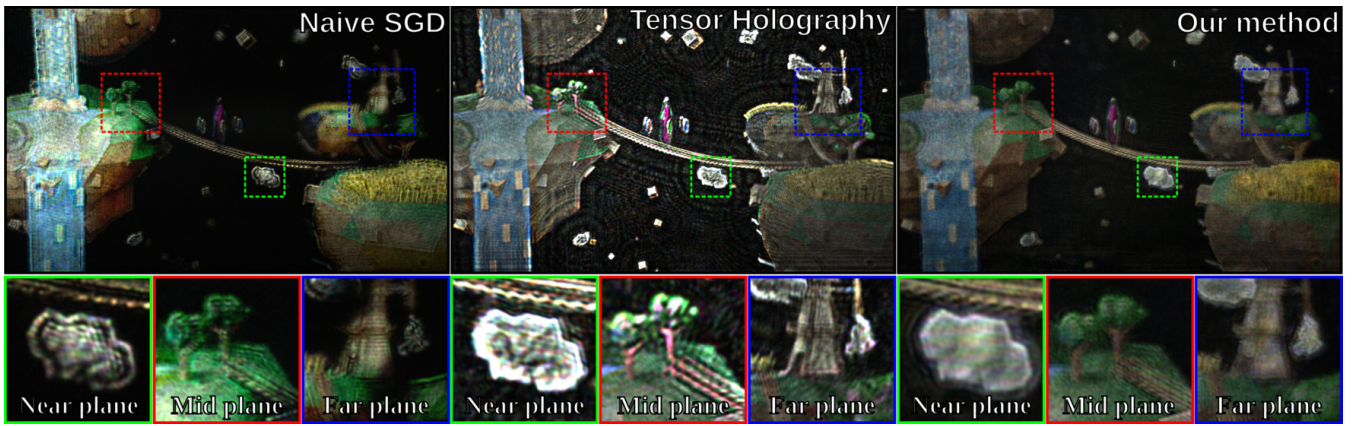

Figure 9: Support for long projection distances. Here, we provide captures from our holographic display at a projection distance of 15 cm. The indicated focuses of near, mid and far correspond to 14.0 cm, 14.5 cm and 15.0 cm, respectively. Our method maintains image quality at these projection distances, unlike the literature's state of the art methods.


\end{document}